\pdfoutput=1
\documentclass{article}
\usepackage{ijcai21}

\usepackage{times}

\usepackage{soul}
\usepackage{url}
\usepackage[hidelinks]{hyperref}
\usepackage[utf8]{inputenc}
\usepackage[small]{caption}
\usepackage{graphicx}
\usepackage{amsmath}
\usepackage{booktabs}
\usepackage{color}
\usepackage{amssymb}
\usepackage{multirow}
\usepackage{multicol}
\usepackage{enumitem}
\usepackage[normalem]{ulem}

\newtheorem{theorem}{Theorem}
\newtheorem{lemma}{Lemma}
\newtheorem{proof}{Proof}
\newtheorem{definition}{Definition}
\newtheorem{corollary}{Corollary}[theorem]
\usepackage{xcolor}
\usepackage{threeparttable}

\definecolor{cite_color}{HTML}{114083}
\definecolor{link_color}{RGB}{153, 0,0}  
\definecolor{url_color}{RGB}{153, 102,  0}
\definecolor{emp_color}{RGB}{0,0,255}
\hypersetup{
 colorlinks,
 citecolor=cite_color,
 linkcolor=link_color,
 urlcolor=url_color}
 
\newcommand{\mbf}[1]{\mathbf{#1}}

\newcommand{\mcal}[1]{\mathcal{#1}}
\def\W{\mathbf{W}}
\def\x{\mathbf{x}}
\def\X{\mathbf{X}}
\def\xx{\times}
\def\R{\mathbb{R}}
\def\V{\mathcal{V}}
\def\E{\mathcal{E}}
\def\Th{\mathbf{\Theta}}

\def\H{\mbf{H}}
\def\D{\mbf{D}}
\def\W{\mbf{W}}

\def\P{\mbf{P}}
\def\Q{\mbf{Q}}

\title{Preventing Over-Smoothing for Hypergraph Neural Networks}
\author{
Guanzi Chen$^1$\footnote{Equal contribuion}\and
Jiying Zhang$^{2*}$\and
Xi Xiao$^2$\and
Yang Li$^1$
\affiliations
$^1$Tsinghua-Berkeley Shenzhen Institute, Tsinghua University\\
$^2$Shenzhen Internationl Graduate School, Tsinghua University\\
\emails
\{cgz20, zhangjiy20\}@mails.tsinghua.edu.cn \and
\{xiaox, yangli\}@sz.tsinghua.edu.cn
}

\begin{document}

\maketitle

\begin{abstract}
In recent years, hypergraph learning has attracted great attention due to its capacity in representing complex and high-order relationships. However, current neural network approaches designed for hypergraphs are mostly shallow, thus limiting their ability to extract information from high-order neighbors. In this paper, we show both theoretically and empirically, that the performance of hypergraph neural networks does not improve as the number of layers increases, which is known as the over-smoothing problem. To avoid this issue, we develop a new deep hypergraph convolutional network called Deep-HGCN, which can maintain the heterogeneity of node representation in deep layers. Specifically, we prove that a $k$-layer  Deep-HGCN simulates a polynomial filter of order $k$ with arbitrary coefficients, which can relieve the problem of over-smoothing. Experimental results on various datasets demonstrate the superior performance of the proposed model compared to the state-of-the-art hypergraph learning approaches.

\end{abstract}
\section{Introduction}

Graphs are ubiquitous and are widely used to represent interactions between entities. \textbf{G}raph \textbf{C}onvolutional \textbf{N}etworks(GCNs) generalizes convolutional neural networks to graph-structure data and have shown superior performance traditional methods in various applications~\cite{GCN,zhang2022fine,chen2022graphtta,chen2021diversified}.  However, graphs can only represent pairwise associations, inadequate in analyzing high-order relationships in the real world. For example, co-authorship may involve more than two authors and communities in social networks often involve more than two person. Naively squeezing such complex relationships into pairwise ones will inevitably lead to unexpected loss of information. 

Hypergraph, a generalization of graph containing edges that can be incident to more than two vertices, serve as a natural tool to model such complex and high order relationships. For example, In a co-citation hypergraph, hypernodes represent papers and hyperedges mean the citation relationships. The ubiquity of complex relationships in the real world naturally encourages the study of hypergraph learning,
including clustering of categorical data~\cite{zhou2007learning}, multi-label classification~\cite{sun2008hypergraph},  image segmentation~\cite{kim2011higher}, image classification~\cite{yu2012adaptive}, mapping users across different social networks~\cite{tan2014mapping}, and so on. 

Furthermore, motivated by the fact that GCNs have been successfully applied to a wide range of applications~\cite{GCN,gcnii}, 
many \textbf{H}ypergraph \textbf{C}onvolutional \textbf{N}erual \textbf{N}etworks(HGCNNs) are proposed to obtain the representation of hypergraph~\cite{zhang2022}. For example, 
HGNN~\cite{hgnn} has been proposed as the first deep learning method on hypergraph structure, employing hypergraph Laplacian to represent hypergraph from spectral perspective. \cite{hypergcn} and \cite{hnhn} proposed a new hypergraph Laplacian to construct HGCNNs, respectively.
Although they have achieved enormous success, most of the current HGCNNs are shadow and the performance degrades when it goes deeper. 
Such a phenomenon also occurs in GCNs and this phenomenon is called over-smoothing~\cite{li2018deeper}. To tackle this problem, many methods have been proposed, such as GCNII~\cite{gcnii}, Scattering GCN~\cite{ScatteringGCN} and DGC~\cite{DGN}.  However, to the best of our knowledge, few work has specially studied the over-smoothing problem on hypergraphs, i.e. whether it exists and if so, how to prevent it. The methods developed for GCNs can not be easily extended to hypergraphs, for GCNs are based on graph normalized Laplacians but not on hypergraph Laplacians.

In this paper, 
on one hand,
we prove that the deep HGCNNs will undergo over-smoothing from the two aspects: stable distribution and Dirichlet energy. Specifically, the vanilla HGCNN(i.e. HGNN~\cite{hgnn}) can be associated with a random walk and eventually converges to the stationary state. And the Dirichlet energy \cite{2020note} of hypernode embeddings will converge to zero, resulting in the loss of discriminative power. 
On the other hand,
we develop a \textbf{Deep} \textbf{H}ypergraph \textbf{C}onvolutional \textbf{N}etwork called Deep-HGCN to alleviate the problem of over-smoothing. Specifically, Deep-HGCN extends two simple but effective techniques: {\it Initial residual} and {\it Identity mapping} \cite{gcnii} to the hypergraph setting.
Initial residual is designed to prevent the convergence of node state by improving the heterogeneity of node representation in deep layers and identity mapping is used to slow down the convergence of Dirichlet energy.
At each layer, initial residual
connects the input layer, while identity mapping adds an identity matrix to the parametric weight matrix. 
Furthermore, 
we prove that a $k$-layer Deep-HGCN model can express a polynomial spectral filter of order $k$ with arbitrary coefficients, which prevents the random walk from converging to  a stationary distribution. 

The main contributions of this paper are summarized as follows:

\begin{enumerate}[leftmargin=*]

\item We prove that HGCNNs have over-smoothing issues from the perspective of random walk and Dirichlet energy. Extensive experimental results also verify this phenomenon.

\item We further propose a truly deep HGCNN called Deep-HGCN,  which adopts two simple yet effective techniques: {\it Initial residual} and {\it Identity mapping}. The fact that a $k$-layer  Deep-HGCN model can express a polynomial spectral filter of order $k$ with arbitrary coefficients proved Deep-HGCN overcomes the over-smoothing problem.

\item We have conducted experiments on various hypernode classification datasets. The experimental results indicated that Deep-HGCN consistently relieves the over-smoothing problem and outperforms state-of-the-art methods.
\end{enumerate}
\section{Related work}
\paragraph{Hypergraph Learning}
Hypergraph learning is first introduced in a seminal work ~\cite{zhou2007learning}, and has been used to encode high-order correlations in many applications, such as image retrieval~\cite{huang2010image}, image segmentation~\cite{kim2011higher}, 3D model classification~\cite{gao2020hypergraph}. Inspired by the superior success of GCNs models, many hypergraph neural network are proposed to obtain the representation of hypergraphs~\cite{li2022a}. HGNN~\cite{hgnn} first employ hypergraph Laplacian to represent hypergraph from spectral perspective. HyperGCN~\cite{hypergcn} trains a GCN for semi-supervised learning on hypergraphs via a new hypergraph Laplacian, which converts a hypergraph to a simple graph. HNHN~\cite{hnhn} designed a convolution network combined both hypernodes and hyperedges features, which based on the hypergraph normalization with cardinality. \cite{zhang2022learnable} propose a learnable hypergraph Laplacian module for updating hypergraph topology during training.

\paragraph{Oversmoothing in GNNs}
~\cite{li2018deeper} first identified the phenomenon that performance will instead degrade when the GCNs models go deeper as over-smoothing. And there are a few other works proposed to addressing oversmoothing. JKNet~\cite{JKNet} proposed Jumping Knowledge Networks, jumping connections into final aggregation mechanism. APPNP~\cite{APPNP} proposed a propagation scheme derived from personalized PageRank. DropEdge~\cite{dropedge} proposed to a randomly drop out certain edges from the input graph. PairNorm~\cite{pairnorm} proposed a novel normalization layer. GCNII~\cite{gcnii} extend vanilla GCN with initial residual and identity mapping.
DGN~\cite{DGN} presented a group normalization layer. ScatteringGCN~\cite{ScatteringGCN} proposed to augment conventional GCNs with geometric scattering transforms and residual convolutions.

\section{Understanding Oversmoothing} \label{sec:3}
In this part, we introduce the hypergraph Laplacian and the extend model HGNN~\cite{hgnn}. Further, we simplified the HGNN to obtain a simple HGCNN called SHGCN via removing the nonlinear transition functions between each layer and only keep the final softmax funtion. Based on HGNN and SHGCN, we conclude that the HGCNNs suffer from over-smoothing problem.
\\
\textbf{Notations.}
Given a hypergraph $\mathcal{G}=(\mathcal{V},\mathcal{E},\mathbf{W})$, $\mathcal{V}$ is the vertex set in which each hypernode $i\in\mathcal V$ is associated with a feature vector $\x_i\in \R^{d}$ where $\X=[\x_1,...,\x_{|\V|}]^T$ denotes the feature matrix, $\mathcal{E}$ is the hyperedge set, and $\mbf{W}$ is the weight matrix of hyperedges.

The structure of hypergraph $\mathcal G$ can be denoted by a incidence matrix $\mathbf{H}\in \mathbb{R}^{|\mathcal{V}|\times |\mathcal{E}|}$ with each entry of $h(v,e)$, which equals 1 when $e$ is incident with $v$ and 0 otherwise. The degree of hypernode and hyperedge are defined as $d(v)=\sum_{e\in \mathcal{E}}w(e)h(v,e)$ and $\delta
(e)=\sum_{v\in\mathcal V}h(v,e)$ which can be denoted by diagonal matrices $\mathbf{D}_v\in \mathbb{R}^{|\mathcal{V}|\times |\mathcal{V}|} $ and $\mathbf{D}_e\in \mathbb{R}^{|\mathcal{E}|\times |\mathcal{E}|}$.

\begin{figure*}[th] 
	    \vspace{-4mm}
	\centering
	\includegraphics[width=1.0\linewidth]{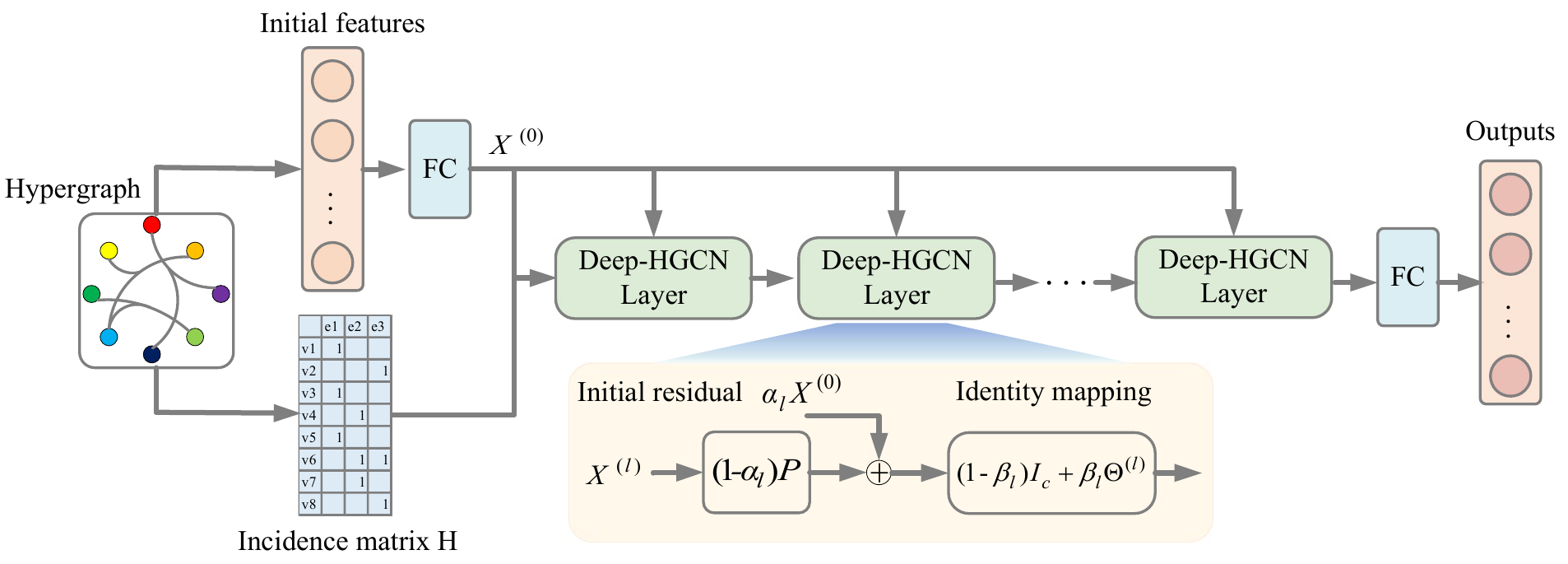}
	\caption{The structure of Deep-HGCN model. FC represents fully connected layer.}
	\label{fig:deep-hgcn}
	    \vspace{-2mm}
\end{figure*}

\subsection{Preliminaries on Hypergraph Convolutions}\label{sec:3.1}
\paragraph{Hypergraph Spectral Convolution}
 ~\cite{zhou2007learning} introduce the normalized hypergraph laplacian  matrix as $\Delta=\mbf{I}-\mbf P$, where $\mbf{P}=\mbf{D}_{v}^{-\frac{1}{2}}\mbf{H}\mbf{W}\mbf{D}_e^{-1}\mbf{H}^T\mbf{D}_{v}^{-\frac{1}{2}}$. $\Delta$ is a symmetric positive semi-definite matrix with spectral decomposition $\Delta=\Phi \Lambda \Phi ^T$. Then, the spectral convolution of a signal $\x=(x_1,\cdots,x_n)$ and a filter $g$ can be denoted as $g*\x=\Phi ((\Phi ^Tg)\odot(\Phi ^T\x))=\Phi g(\Lambda)\Phi ^T \x$.
~\cite{hgnn} indicate that the filter $g(\Lambda)$ can be approximated by a $K$-th order polynomial of $\Lambda$, and  
the hypergraph convolution can be further approximated by the $K$-th order polynomial of Laplacians
\begin{align}
     \Phi g_\theta \Phi^T\x \approx \Phi \left(\sum_{k=0}^K \theta_k \Lambda^k\right)\Phi ^T \x  =\left(\sum_{k=0}^K\theta_k\Delta^{k}\right)\x,
\end{align}
where $\theta\in \R^{K+1}$.
HGNN~\cite{hgnn} sets $K=1$, $\theta_0=-\frac{1}{2}\theta$, $\theta_1=\frac{1}{2}\theta \mbf P$ to derive $g*\x=\theta \mbf P \x$, and finally bulid a Hypergraph Convolutional Layer
\begin{equation}
    \begin{aligned}
    \mbf{X}^{(l+1)}=\sigma(\mbf P\mbf{X}^{(l)}\Th ^{(l)}). \label{Eq:hgnn}
   \end{aligned}
\end{equation}
where $\sigma$ denotes the nonlinear activation function.
\paragraph{Simple Hypergraph Convolutional Network} In order to better analyze the performance of the multi-layer hypergraph convolutional networks, referring to SGC~\cite{wu2019simplifying}, we remove the non-linear activation function betwen each layer in Eq.\eqref{Eq:hgnn} and add the softmax in the final layer to obtain \textbf{S}imple \textbf{H}yper\textbf{G}raph \textbf{C}onvolutional \textbf{N}etwork called SHGCN, of which each layer can be denoted as:
\begin{align}
    \X^{(l+1)}= \mbf P\X^{(l)}\Th^{(l)} 
\end{align}
and the output can be formulate as:
\begin{align}
    Y=softmax(\mbf P^{l}\X^{(0)}\Th) \label{Eq:SHGCN},
\end{align}
where $\Th=\prod_{k=0}^{l}\Th^{(l)}$ is a learnable parameter matrix and $\X^{(0)}=\X$, representing the input features. Actually a $K$-layer SHGCN simulates a polynomial filter of order $K$ with fixed coefficients $\mbf{\theta}=[\theta_0,...,\theta_K]$ since $\mbf P^{K}\x$=$(\mbf{I}-\Delta)^{K}\x$= $\left(\sum_{k=0}^{K}\theta_l\Delta^{k}\right)\x$. The biggest difference between SHGCN and HGNN is that the removed activation function, which is usually taken ReLU and can not essentially affect the process of convolution. So the over-smoothing problem of HGNN can explore by SHGCN directly.
Then we will show that the fixed filter limits the ability of expressive power and lead to over-smoothing.

\subsection{ Studying Over-smoothing with Random Walk} \label{sec:3.2}
According to ~\cite{zhou2007learning}, each hypergraph can be associated with a natural random walk with transition matrix $\mbf T=\mbf{D}_{v}^{-1}\mbf{H}\mbf{W}\mbf{D}_e^{-1}\mbf{H}^T$. The stationary distribution of the random walk is 
\begin{align}
    \mbf{\pi}=\mbf{1}^T\mbf{D}_v /\operatorname{vol}\mcal{V}   \label{eq:sd1}
\end{align}
where $\mbf 1\in \R^{ |V|\xx 1}$ is an all-one vector, $\pi\in \mathbb{R}^{1\xx |\V|}$ is the probability vector, and  $\operatorname{vol}\mcal{V}$ is the sum of the degrees of the vertices in $\mcal{V}$. Then we can derive the stationary distribution w.r.t. $\mbf P$.
\begin{lemma}  
\label{yl:sd1}
    If $\mathcal{G}=(\mathcal{V},\mathcal{E},\mbf{W})$ is a hypergraph, then
    the stationary distribution of a random walk with transition matrix $\mbf P=\mbf{D}_{v}^{-\frac{1}{2}}\mbf{H}\mbf{W}\mbf{D}_e^{-1}\mbf{H}^T\mbf{D}_{v}^{-\frac{1}{2}} $ on $\mcal G$ is
    \begin{align}
        \mbf{\Tilde{\pi}}=\mbf{1}^T\mbf{D}_v^{\frac{1}{2}}/\operatorname{vol}\mcal{V}    
    \end{align}
   
\end{lemma}

The proof of Lemma \ref{yl:sd1} can be found in supplementary materials.
Next, we build the connection between \textit{random walk} and \textit{Laplacian smoothing} to further explore the hypergraph convolution. Considering a one-layer SHGCN. It actually contains two steps.\\
1) Generating a new feature matrix Y from X by applying the hypergraph convolution:
\begin{align}
    Y = \mbf P \X \label{conv}
\end{align}
2) Feeding the new feature matrix $Y$ to a fully connected
layer. \\
The hypergraph convolution Eq.\eqref{conv} actually aggregates the message of neighbor hypernodes and reduces the heterogeneity of hypernodes representation, which can be regraded as a special form of Laplacian smoothing~\cite{li2018deeper}. Then we show that the Laplacian smoothing lead to the problem of over-smoothing.
Let $\x_{*j}\in\R^{|\V|}$ denote the $j$-th column of $\X$ base on lemma \ref{yl:sd1}, We derive the following theorem.
\begin{theorem}
\label{dl:1}
Repeatedly applying Laplacian smoothing on $\x_{*j}$, we have
    \begin{align}
        \lim_{l\to\infty}\mbf{P}^{l}\x_{*j}=\pi_j^T\label{Eq:jixiang}
    \end{align}
    where $\pi_j=\left(\sum_{i=1}^{|\V|}\x_{ij}\right)\Tilde{\pi} \in \R^{1\xx |\V|}$.
    
\end{theorem}

\noindent The proof of Theorem \ref{dl:1} is provided in supplementary materials.
The Theorem \ref{dl:1} means that the repeatedly apply Laplacian smoothing on $\x_{*j}$  will converge to a stationary point, which only depends on the sum of initial features $\sum_{i=1}^{|\V|}\x_{ij}$ and the stationary distribution $\Tilde{\pi}$ that is only related to hypergraph structure(i.e.,degree). That is to say, the $l$-th representation $\mbf P^{l} \X$ of SHGCN converge to the matrix $\left[\sum_{i=1}^{|\V|}\x_{i1}\tilde{\pi}^T,...,\sum_{i=1}^{|\V|}\x_{id}\tilde{\pi}^T\right]\in\R^{|\V|\xx d}$ which only 
contains the information of initial features and degree of each nodes leads to over-smoothing. what's worse, the information of initial features just the sum of column of the $\X$, which does not contain any hypernode feature. Such convergence suggests that the $K$-order polynomial filter $\mbf P^K \X$ with fixed coefficients cause the problem of over-smoothing.

\subsection{ Studying Over-smoothing with Dirichlet energy}\label{sec:3.3}
\cite{2020note} suggests that expressiveness of GCNs can be measured by the Dirichlet energy and the over-smoothing problem is cause by the fact that the Dirichlet energy of embeddings converges to zero. So in this part, We will investigate the expressive power of HGCNNs via their Dirichlet Energy as the layer size tends to infinity.

\begin{figure}
    \centering
    \includegraphics[width=8cm]{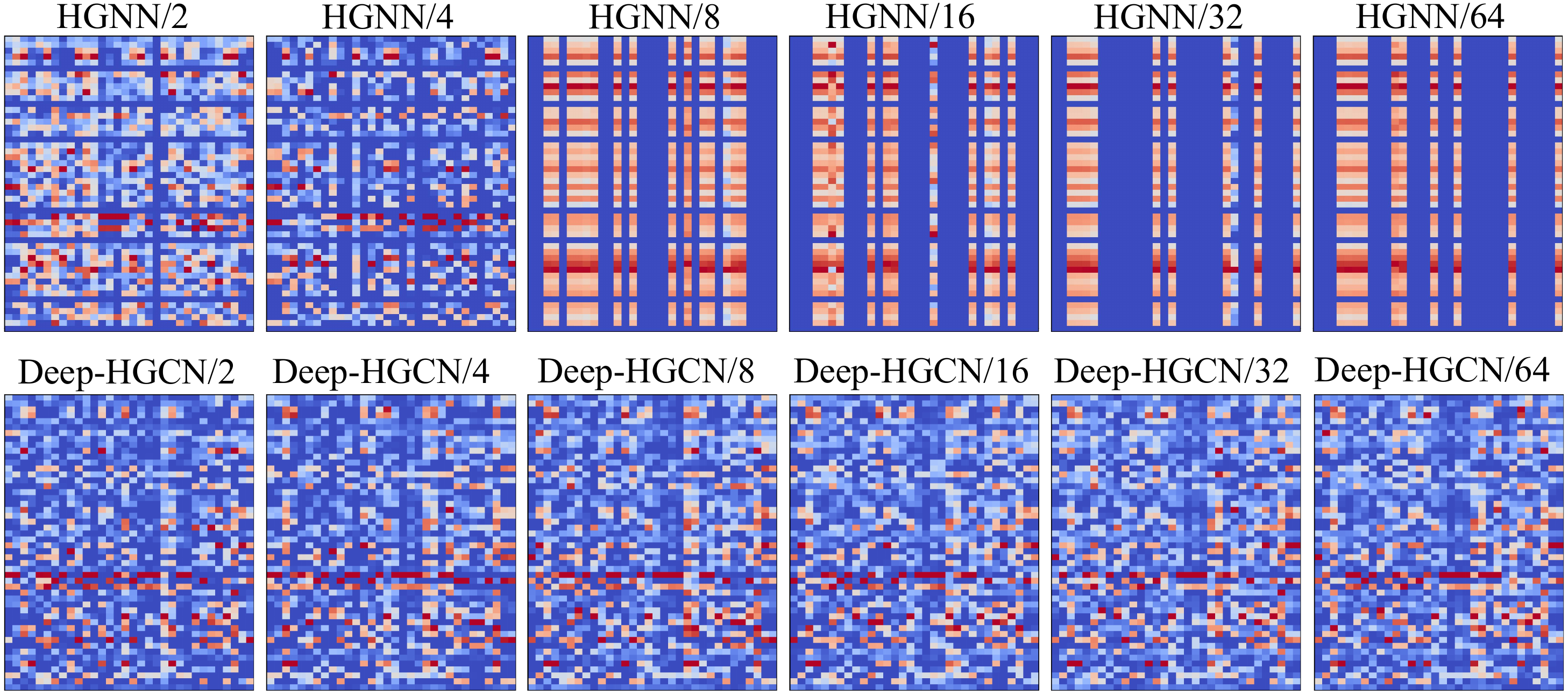}
    \caption{Visualization of hypernode embeddings on co-citation Cora. A node is represented by a row of the matrix. As the number of layers increase,  the node embeddings of HGNN tend to perform high homogeneity, indicating the serious over-smoothing phenomenon. The Deep-HGCN instead preserves much more diversified representation patterns. }
    \label{fig:hotmap}
        \vspace{-2mm}
\end{figure}

\begin{definition}\label{dy:hypergraph_energy}
    Dirichlet energy $E(f)$ of a scalar function $f\in \R^{d\xx1}$ on the hypergraph $\mathcal{G}=(\mathcal{V},\mathcal{E},\mbf{W})$ is defined as:
    \begin{align}
    		E(f)=f^T\Delta f
    		&= \frac{1}{2}\sum_{e\in\E}\sum_{\{u,v\}\in \V}
    		\frac{w(e)h(u,e)h(v,e)}{\delta(e)} \notag\\
    		&\left(\frac{f(u)}{\sqrt{d(u)}}-\frac{f(v)}{\sqrt{d(v)}}\right)^2,
    		\label{eq:DE}
    \end{align}
\end{definition} 
For the features matrix $\X=[\x_1,...,\x_{|\V|}]^T$, the Dirichlet energy is defined as $ E(\X)=tr(\X^T\Delta\X)$. 
Recall Eq.\eqref{Eq:hgnn}, the representation of $l$-th layer in HGNN~\cite{hgnn} can be denoted as:
\begin{align}
    \X^{(l)}=\sigma(\sigma(\cdots \sigma(\P\X) \Th^{(0)} ) \Th^{(1)})\cdots \Th^{(l-1)}) \label{Eq:expand}
\end{align}
Next we study the influence of $\P$, $\Th_l$ and the activation function $\sigma$ on the Dirichlet energy respectively.
\begin{lemma}\label{yl:2}
The hypergraph energy is same as Definition \ref{dy:hypergraph_energy}, we have \\
    (1) $E(\P\X)\leq (1-\lambda)^2E(\X)$, where $\lambda$ is the minimum non-zero eigenvalue of $\Delta$. \\
    (2) $ E(\X \Th) $ $\leq \| \Th^T \|_{2}^{2}E(\X) $, where $\| \cdot \|_2$ represents the maximum singular value of the matrix. \\
    (3) $E(\sigma(\X))\leq E(\X)$, where $\sigma$ is ReLU or Leaky-ReLU activation function.
\end{lemma}
The above Lemma \ref{yl:2} can be directly proved by the definition of  hypergraph energy. More detail about the proof can be see in the supplementary materials. Thus, we denote $s_l:=\prod_{k=0}^{l-1}\|\Th^{(k)}\|_2^2$ and $\Bar{\lambda}:=(1-\lambda)^2$ and deduce the main theorem as follow.

\begin{theorem}
    For any $l\ge1$, we have $E(\X^{(l)})\leq s_l\Bar{\lambda}E(\X^{(l-1)}) $. \label{dl:2}
\end{theorem}
Based on Eq.\eqref{Eq:expand} and Lemma \ref{yl:2}, the theorem can be proved. The detail of proof can be found in the supplementary materials.
\begin{corollary}
    \label{tl:2.1}
    Let $s:=\operatorname{max}_{1\leq k\leq l}s_k$, we have $E(\X^{(l)})\leq s^l\Bar{\lambda}E(\X)$. In particular, 
    $E(\X^{(l)})$ exponentially converges to zero when $s\Bar{\lambda}< 1$.
\end{corollary}
From the Corollary \ref{tl:2.1}, we know the Dirichlet energy of the embedding of HGNN will converge to zero if $s<1/\Bar{\lambda}$, which means that the HGCNNs will suffer from over-smoothing. And the result also reveals that  $s$ is a key point that we can operate to improve the expressive power of a multi-layer model.

\section{Deep-HGCN: Tackling Over-Smoothing}
Recall to section~\ref{sec:3}, we get two main reasons that cause the problem of hypergraph convolution: 1) The fixed coefficient filter; 2) The diminishing Dirichlet energy. So in the part, we mainly focus on these two problems to design the multi-layer model.
\subsection{Deep-HGCN Model}
In order to improve the expressive power of deep HGCNNs, we can make the fixed coefficients flexible, which is a key point to prevent the over-smoothing~\cite{gcnii}. Here, we adopts two techniques: \textit{Initial residual} and \textit{Identity mapping} to obtain the truly deep network Deep-HGCN. The two techniques are proved effective in preventing over-smoothing in GNNs~\cite{gcnii}. Thus, the $l$-th layer of Deep-HGCN can be denote as 
\begin{align}
    \mbf{X}^{(l+1)}=\sigma\left(\left((1-\alpha_l)\mbf{P}\mbf{X}^{(l)}+\alpha_l\mbf{X}^{(0)}\right)\Th^{(l)}_{I}\right),\label{Eq:deep-hgcn}
\end{align}
where  $\alpha_l$ and $\beta_l$ are two hyper-parameters, $\Th^{(l)}_{I}=(1-\beta_l)\mbf{I}_{c}+\beta_l\mbf{\Th}^{(l)}\in \R^{c\xx c}$ is the learnable weight matrix with identity mapping. As for $\X^{(0)}$, we can set to the linear transform of input feature $\X\in\R^{|\V|\xx d}$ if $d$ is large $(h\leq d)$. Recall that $\mbf P= \D_v^{-\frac{1}{2}} \H\W \D_v^{-\frac{1}{2}} \H^T \D_v^{-\frac{1}{2}}$ is the hypergraph convolution matrix. Comparing Eq.\eqref{Eq:hgnn} with Eq.\eqref{Eq:deep-hgcn}, we notice two differences: 1) Eq.\eqref{Eq:deep-hgcn} add a identity matrix $\mbf I_{c}$ to the weight $\Th$; 2) Eq.\eqref{Eq:deep-hgcn} add the first layer representation $\X^{(0)}$ to the current layer Laplacian smoothing representation $\mbf P\X^{(l)}$.

\paragraph{Initial Residual connection} \cite{bai2019hypergraph} proposes the Skip Connection to connect the smoothed feature $\Tilde{\mbf P}\X^{(l)}$ and $\X^{(l)}$. However, they also show the performance deterioration as the model become deeper but do not give the reason. 
We start from the perspective of the stationary distribution of random walk to reconsider the Skip Connection. Based on the analysis over-smoothing in section \ref{sec:3.2}, 
we know over-smoothing washes away the signal from all the features,
making them indistinguishable. So we add the representation of the first layer to each layer for compensating the heterogeneity of the hypernodes.
The hyper-parameters $\alpha_l$ indicates how much the initial features information that each layer can carry.
Despite we stack many layers, it can receive at least $\alpha_l$ proportion message from the input layer, which ensure the performance at least one layer of the model.

\paragraph{Identity mapping} 
It seems that we can alleviate the over-smoothing by adding the initial residuals in each layers, however, in GNNs, APPNP~\cite{APPNP} adopt the Initial Residual connection only obtain a shallow model, which still undergo the over-smoothing problem. And our experiments also show that the same is true for HGCNNs(Figure \ref{fig:ablation}). 
In order to improve the performance of the Deep-HGCN when the number of layers increases, we add the the identity matrix to weight matrix. The motivation as follow: 

\begin{figure}[thbp]
\definecolor{RRed}{RGB}{223,25,27}
\definecolor{BBlue}{RGB}{54,123,180}
\definecolor{GGreen}{RGB}{75,171,72}
\definecolor{OOrange}{RGB}{249,124,0}
	\centering
	\includegraphics[width=0.45\linewidth]{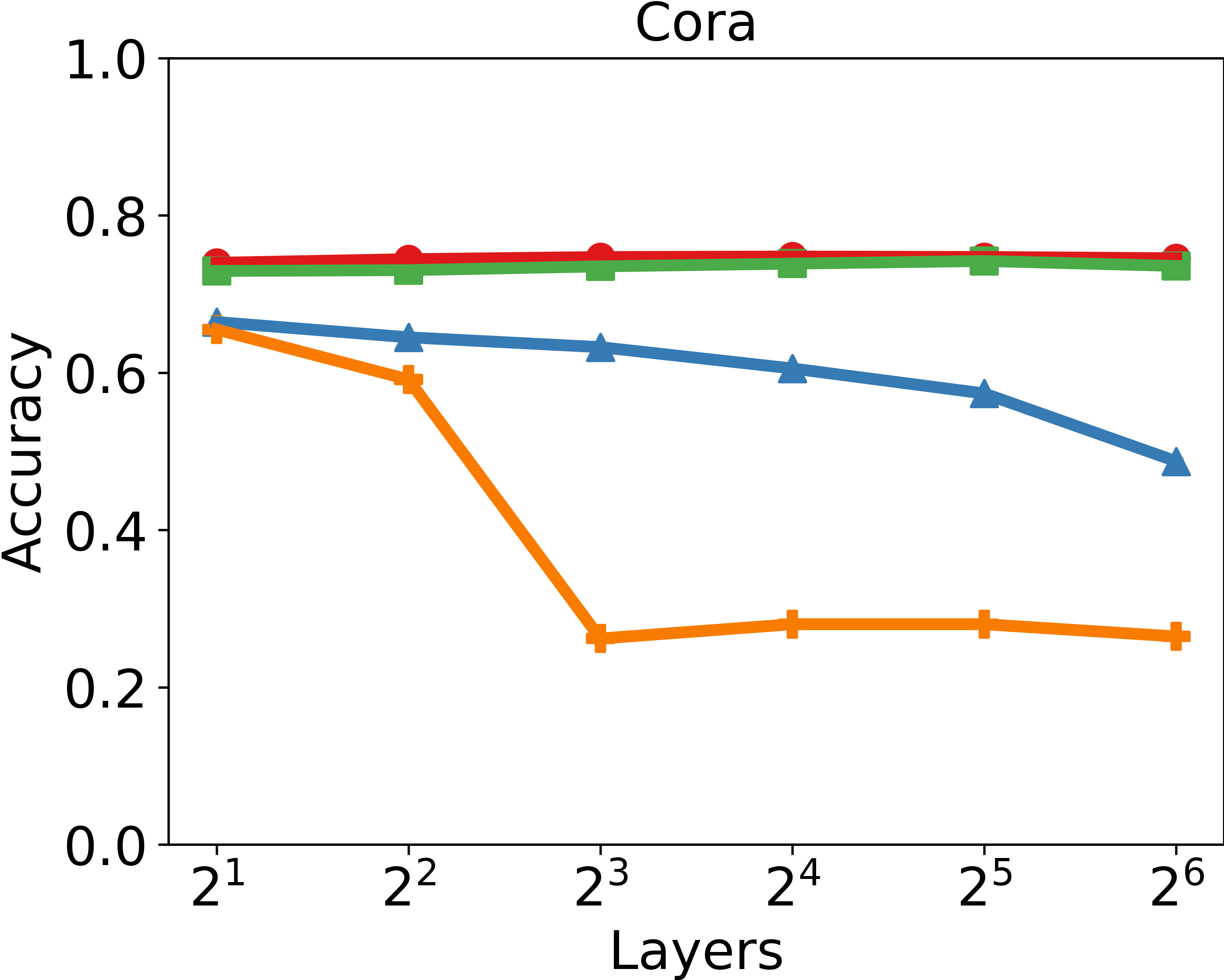}
	\quad
	\includegraphics[width=0.45\linewidth]{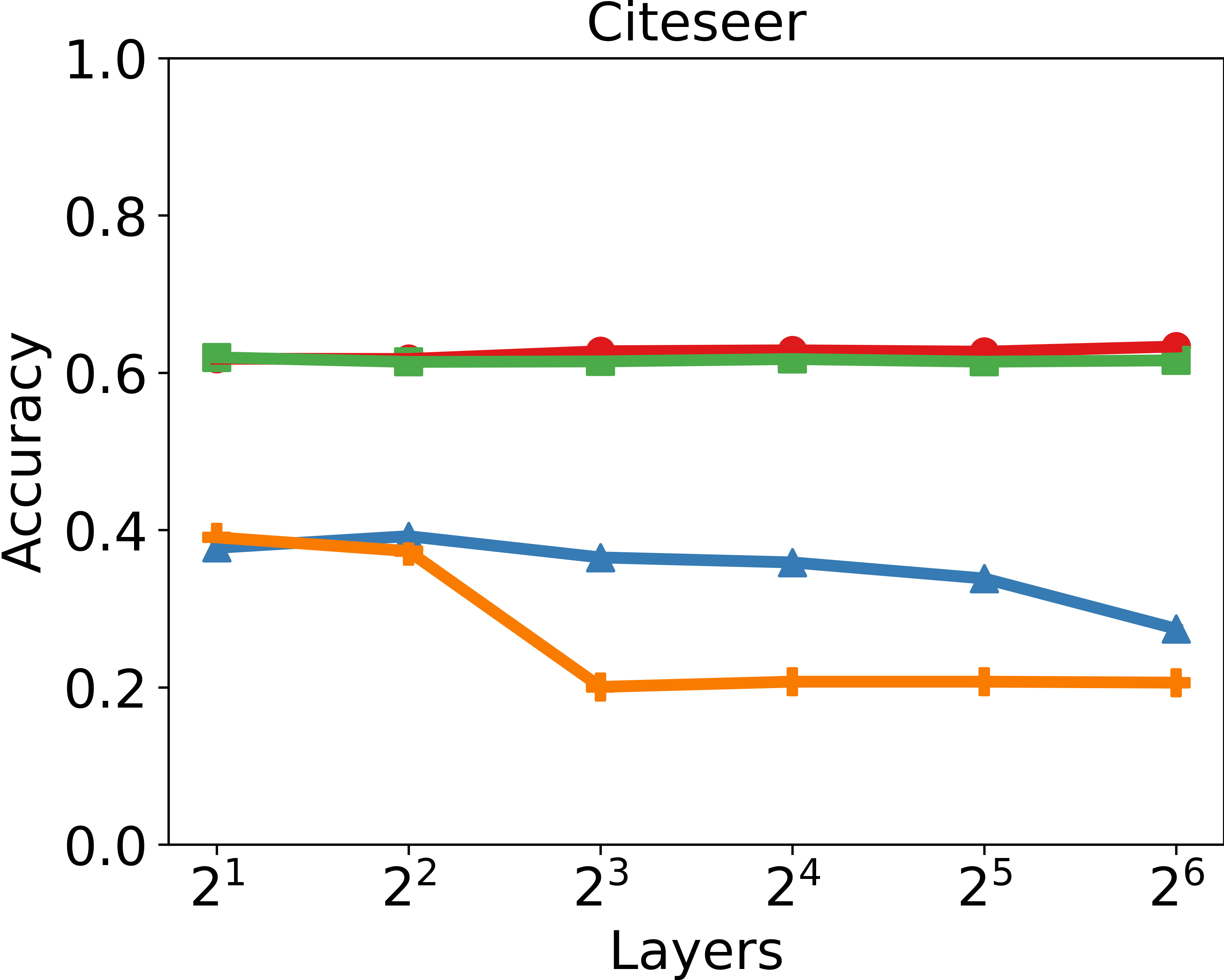} 
	\caption{Ablation study on initial residual and identity mapping.\textcolor{RRed}{Red}: our proposed Deep-HGCN, \textcolor{GGreen}{Green}: the variant without Identity Mapping, \textcolor{BBlue}{Blue}: the variant without Initial Residual, and \textcolor{OOrange}{Orange}: the variant without Initial Residual and Identity Mapping.(HGNN) }
	\label{fig:ablation}
	    \vspace{-0mm}
\end{figure}

\begin{itemize}[leftmargin=*]
    \item \textbf{Slow down the Dirichlet energy convergence to zero.} According to the analysis of Dirichlet energy in section~\ref{sec:3.3}, we know that the rate of convergence of energy depends on the $s^K$, where $ s=\operatorname{sup}_{1\leq l\leq K}s_l$. 
    Replacing the $\Th^{(l)}$ by $\Th^{(l)}_{I}$ would make the largest singular value of the $\Th^{(l)}$ (i.e. $s_l$) close to $1$ if we apply regularization on $\Th^{(l)}$ and force the norm of it to small. Consequently, The trend of $s^K$ tending to zero will slow down, so that the model can alleviate the energy reduction.
    
    \item \textbf{Useful in designing deep network.} 
    ResNet~\cite{ResNet} adds the identity mapping to ensure the training error of a deeper model no more than its shallower counter-part. So similar to the motivation of it, we expect to achieve that the performance of deeper model at least same to the shallow model.
    
    \item \textbf{Effective in solving over-smoothing problems in GCNs.} 
    GCNII~\cite{gcnii} add the identity matrix to weight matrix and get a truly deep GCN and indicate the technique is effective in the semi-supervised task. Notice the hypergraph can be seem as the generalization of graph, we apply identity mapping to Deep-HGCN and expect it can work on hypergraph convolution.
\end{itemize}

To ensure the retard the Dirichlet energy deteriorate, we should set the $\beta_l$ decreases adaptively as the number of layers increases. Here, we adopt the setting $\beta_l=log(\frac{\lambda}{l}+1)$ as suggested in \cite{gcnii}.

\subsection{Spectral Analysis} \label{sec:4.2}

Recall the analysis in section \ref{sec:3.2}, a $K$-layer SHGCN simulates a polynomial filter $\left(\sum_{l=0}^{K}\theta_l\Delta^{l}\right)\x$ of order $K$ with fixed coefficients $\theta$ on the hypergraph spectral domain of $\mathcal{G}$, and the fixed coefficients limits the ability of learning the distinguishable hypernode representations of a multi-layer HGCNNs and thus cause over-smoothing. Now, we prove that a $K-$ layer Deep-HGCN simulates a polynomial of order $k$ with arbitrary coefficients.

 \begin{table}[htbp]
\def\p{$\pm$} 
\centering

\caption{Summary of classification accuracy(\%) results with various depths.We report mean test accuracy over 10 train-test splits. The co-a denotes co-authorship and co-c denotes co-citation}
    \vspace{0mm}
\scalebox{0.88}{
\begin{tabular}{p{0.75cm}p{1.9cm}|p{0.6cm}p{0.6cm}p{0.6cm}p{0.6cm}p{0.6cm}p{0.6cm}}
\toprule
\multirow{2}{*}{Dataset} & \multirow{2}{*}{Method} & \multicolumn{6}{c}{Layers}  \\
& & 2     & 4     & 8     & 16    & 32    & 64       \\

\midrule
\multirow{4}{*}{\begin{tabular}[c]{@{}l@{}}Cora~\\(co-a)\end{tabular}} 
& MLP & 50.63 & 51.03 & 50.42 & \textbf{51.34} & 50.74& 50.81    \\
& HyperGCN  & \textbf{52.82} & 50.71 & 23.84  &23.25   &23.13   & 23.08      \\
& HGNN &\textbf{66.36} & 59.47 & 25.69 & 26.27 & 25.26& 27.34     \\
& Deep-HGCN & 73.53 & 74.08 & 74.52 & 74.50  & 74.91  &\textbf{75.08}    \\
\midrule
\multirow{4}{*}{\begin{tabular}[c]{@{}l@{}}DBLP~\\(co-a)\end{tabular}}   
& MLP & 77.34 & 77.31 & \textbf{77.41} & 77.32 &77.34 & OOM \\
& HyperGCN & \textbf{70.11} & 58.72 & 22.87 & 21.73  &22.18   & OOM      \\
& HGNN & \textbf{88.22}& 85.30 & 26.30 & 25.69 &26.57 & OOM  \\
& Deep-HGCN& 89.00 & 89.30  & \textbf{89.33} & 89.14 & 88.80  &OOM  \\
\midrule
\multirow{4}{*}{\begin{tabular}[c]{@{}l@{}}Cora~\\(co-c)\end{tabular}}     
& MLP & 50.63 & 51.03 & 50.42 & \textbf{51.34} & 50.74& 50.81
   \\
& HyperGCN & 48.38 & \textbf{50.75} & 26.37  & 26.36& 23.29 &19.63  \\
& HGNN & \textbf{48.06}& 41.05& 18.24 & 17.54 & 17.92 & 22.81    \\
& Deep-HGCN & 66.90 & 68.12  & 68.33 & 68.18  & 69.00 & \textbf{69.28 }\\
\midrule
\multirow{4}{*}{\begin{tabular}[c]{@{}l@{}}Pubmed~\\(co-c)\end{tabular}}   
& MLP & 71.46 & 71.46 &\textbf{71.46 }& 71.46 & 71.45 & 71.45    \\
& HyperGCN & 56.93 & \textbf{61.92} & 35.21& 32.19  & 35.86  &    36.00   \\
& HGNN & \textbf{42.72} & 39.52 & 36.32 & 36.43 & 36.06 & 36.37    \\
& Deep-HGCN & 74.08 & 74.42 & 74.100 & 74.51& \textbf{74.68} & 74.63\\
\midrule
\multirow{4}{*}{\begin{tabular}[c]{@{}l@{}}Citeseer~\\(co-c)\end{tabular}} 
& MLP & 52.00 & 51.92 & 51.75 & 51.75 & 51.65 &\textbf{52.00}    \\
& HyperGCN & 50.26  & \textbf{57.09} & 21.83 & 21.89  &21.67  & 21.69     \\
& HGNN & \textbf{39.50 }& 37.06& 19.68 & 20.32 & 19.22 & 20.42    \\
& Deep-HGCN & 61.00 & 60.90& 61.85  & 62.10  &  61.69 &  \textbf{62.51}   \\
\bottomrule

\end{tabular}
}
    \vspace{0mm}
\label{tab:depth_accuracy}
\end{table}

\begin{theorem}\label{dl:3}
  Suppose $\x$ is a graph signal in the $\mathcal G$, then 
  a $K$-layer Deep-HGCN can express a $K$ order polynomial filter $\left(\sum_{k=0}^{K} \theta_{k} \Delta^{k}\right)\x$ with arbitrary coefficients $\theta$.
\end{theorem}
The proof of Theorem \ref{dl:3} can be seen in supplementary materials.
The theorem indicate the deep-HGCN replaces the fixed coefficients in SHGCN with the arbitrary coefficients, which can prevent the convergence of  $\P^{K}\x$. Recall the section \ref{sec:3.2}, we know the over-smoothing mainly cause by the fact that the node representation $\P^l\x$  in deep layer converges to a distribution that is independent of input node features. However, as suggest in Theorem \ref{dl:3}, our model can carry the topology and initial node features information of the hypergraph to the deep, which not only enhance the heterogeneity of the node but also indeed relieve the problem of over-smoothing. 

\begin{table*}[thbp]
\def\p{$\pm$} 
\centering
\setlength\tabcolsep{12pt} 
\caption{Summary of classificaiton accuracy(\%) results. The number in parentheses corresponds to the number of layers of the model. We report the average test accuracy and its standard deviation over 10 train-test splits.(OOM: our of memory)}
    \vspace{0mm}
\scalebox{0.88}{
    \begin{tabular}{c|c|c|c|c|c}
        \toprule
              \multicolumn{1}{c}{Dataset} &\multicolumn{1}{c}{\begin{tabular}[c]{@{}c@{}}Cora\\ (co-authorship)\end{tabular}}  & \multicolumn{1}{c}{\begin{tabular}[c]{@{}c@{}}DBLP\\ (co-authorship)\end{tabular}} & \multicolumn{1}{c}{\begin{tabular}[c]{@{}c@{}}Cora\\ (co-citation)\end{tabular}} & \multicolumn{1}{c}{\begin{tabular}[c]{@{}c@{}}Pubmed\\ (co-citation)\end{tabular}} & \multicolumn{1}{c}{\begin{tabular}[c]{@{}c@{}}Citeseer\\ (co-citation)\end{tabular}} \\
        \midrule
         MLP+HLR &  59.8\p4.7 & 63.6\p4.7 & 61.0\p4.1 & 64.7\p3.1 & 56.1\p2.6 \\
             FastHyperGCN &  61.1\p8.2 & 68.1\p9.6 & 61.3\p10.3 & 65.7\p11.1 & 56.2\p8.1 \\
             HyperGCN &  63.9\p7.3 & 70.9\p8.3 & 62.5\p9.7& 68.3\p9.5 & 57.3\p7.3 \\
            HGNN &  63.2\p3.1 & 68.1\p9.6 & 70.9\p2.9& 66.8\p3.7 & 56.7\p3.8 \\
             HNHN~\cite{hnhn} &  64.0\p 2.4 & 84.4\p 0.3& 41.6\p 3.1 & 41.9\p4.7 & 33.6\p 2.1  \\
             HGAT~\cite{HyperGAT} &  65.4$\pm$1.5 & OOM &52.2$\pm$3.5 & 46.3$\pm$0.5 & 38.3$\pm$1.5 \\
        \midrule
             Deep-HGCN(ours) & \textbf{75.08\p1.0(64)} & \textbf{89.33\p0.3(8)} & \textbf{69.28\p1.2(64)} & \textbf{74.68\p0.9(32)} & \textbf{62.51\p1.4(64)}\\
        \bottomrule
    \end{tabular}
    }
    \vspace{0mm}
    \label{tab:sota_accuray}    
\end{table*}

\section{Experiments}
In this section, we evaluate the performance of Deep-HGCN against the state-of-the-art hypergraph convolutional neural network models two classification task: hypernode classification on co-authorship and co-citation networks and visual object recognition.

\subsection{Hypernode classification}
\paragraph{Datasets and Experiment settings} In this experiment, the task is semi-supervised node classification. We use the datasets provided by~\cite{hypergcn} which include three co-citation network datasets: Cora, Pubmed and Citeseer,  and two co-authorship network datasets: Cora, DBLP. We take the same train-test split as the realeased github\footnote{https://github.com/malllabiisc/HyperGCN}(i.e.10 different train-test splits), which different from \cite{hypergcn} reported on paper. 
Statistics of the datasets are summarized in Table \ref{tab:dataset_hypergcn}, (refer to Appendix). 

We use the Adam SGD optimizer to train the model for 300 epochs in total, with a learning rate of $0.01$ and early stopping with a patience of $100$ epochs. 
The feature dimension of the hidden layer is set as 32 and the dropout rate is set as 0.5.  Experiments are done on a GeForce RTX 2080 Ti GPU and the results of baseline(MLP, HyperGCN, HGNN, HNHN, HGAT) are reproduced by their release Codes whose hyper-parameters follow the corresponding papers.We have used grid search to tune hyper-parameters $\alpha$ and $\beta$, the search range is provided in Table~\ref{tab:hyper_parameter}in appendix. 

\paragraph{Comparison with SOTA} The results are shown in Table ~\ref{tab:sota_accuray}.
We can observed that the Deep-HGCN outperform the baselines by a large margin in most cases. The main reason why the results of HyperGCN is different from the corresponding paper is that \cite{hypergcn} reported the results on 100 train-test split datasets.

\begin{figure}[htbp]
    \centering
    \includegraphics[width=8cm]{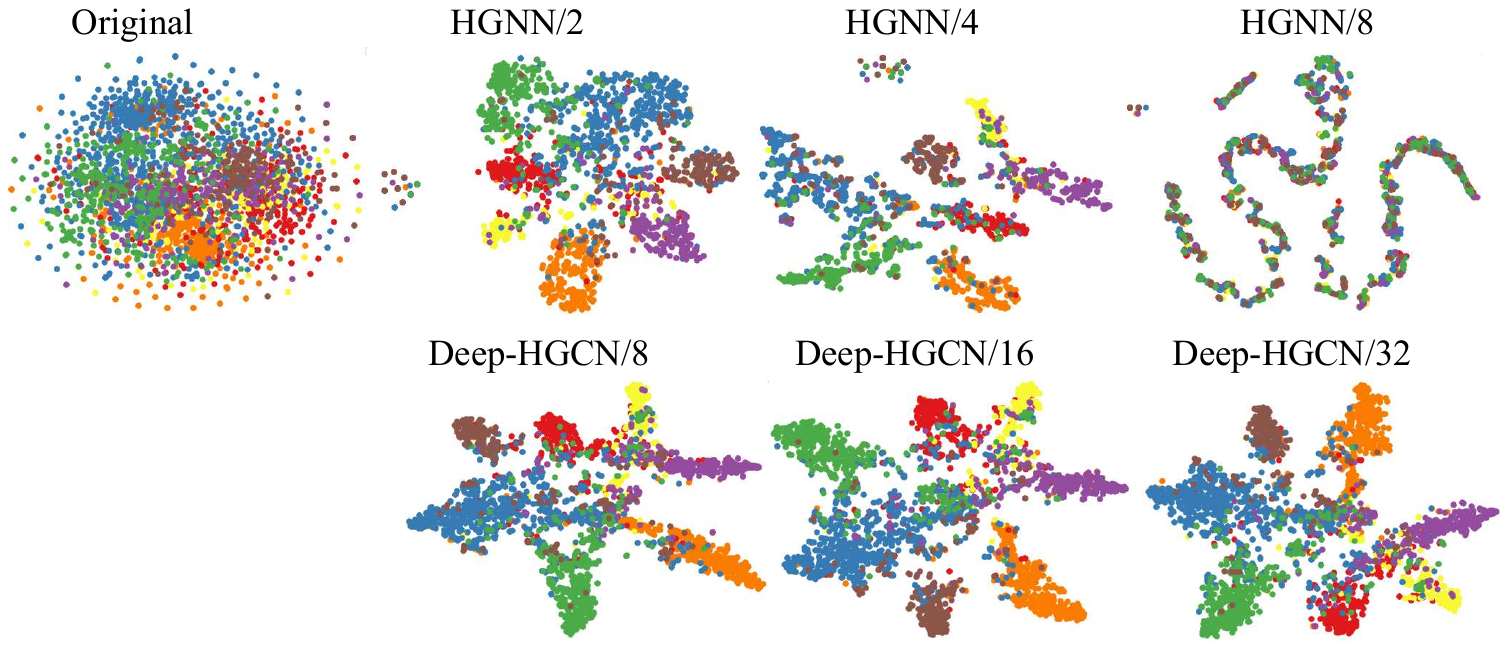}
   \caption{t-SNE visualization of the hypernode representations on Cora. Different colors represent different node classes. %
   As the number of layers increases, the embeddings of different classes on HGNN also tend to be similar and lead to indistinguishable features. Instead, our model can get a more easily distinguishable node embedding.}
    \label{fig:tsne}
    \vspace{0mm}
\end{figure}

\paragraph{Oversmoothing Analysis}Table \ref{tab:depth_accuracy} summaries the results with various depth on the five datasets.  It is observed that HGNN and HyperGCN most achieve their best accuracy with 2 or 4 layers, and the accuracy drop rapidly when the number of layers exceeds 4, suggesting that they suffer from severe over-smoothing. Instead, our proposed Deep-HGCN achieve the best accuracy at layer 64, 8, 16, 64, 32 and 64 respectively. It is noted that the performance of the Deep-HGCN consistently improves as we increase the number of layers, which verifies our analysis that our model can maintain the heterogeneity of node representation in deep layers.

We also draw t-SNE and feature maps for hypernode representations. Figure \ref{fig:tsne} is the t-SNE visualization of learned node representations. Figure \ref{fig:hotmap} is the feature map of learned embedding of the first 50 hypernodes. Overall, the results suggest that HGCNNs suffer from severe over-smoothing problems, while Deep-HGCN can effectively relieve the over-smoothing problem and extend the HGNN into a truly deep model. 

\paragraph{Ablation study} 
 In order to further explore the two techniques: \textit{Initial residual} and \textit{identity mapping}, we do the ablation study of them. As the Figure~\ref{fig:ablation} show, compare to the HGNN, just adopting the identity mapping can effective alleviate the problem of over-smoothing and just applying the initial residual can greatly alleviate the over-smoothing. Using  both of them at the same time increase the performance of the model to a greater extent. The result suggest that both techniques are need to overcome the problem of over-smoothing.

\subsection{Visual object recognition}
\textbf{Datasets and experimental settings} In this experiment, we compare with HGNN~ on two public objects: the Princeton ModelNet40 dataset and the National Taiwan University 3D model dataset NTU2012, where the hypergraph structure construction are as described in HGNN~\cite{hgnn}. We use the same hype-rparameters as \cite{hgnn}

\textbf{Results} The results on visual object recognition task are shown in Table \ref{tab:object_accuracy}. As shown in the results, the proposed Deep-HGCN can achieves much better performance compared with HGNN~\cite{hgnn} on this two datasets. Detailed statistical information can refer to Table~\ref{tab:dataset_object} in Appendix.

\begin{table}[htbp]
\def\p{$\pm$} 
\centering
    \vspace{-0mm}
\caption{Summary of classification accuracy(\%) results with various depths on object datasets.}
    \vspace{-2mm}
\scalebox{0.9}{
\begin{tabular}{p{0.5cm}c|p{0.6cm}p{0.6cm}p{0.6cm}p{0.6cm}p{0.6cm}p{0.6cm}}
\toprule
\multirow{2}{*}{Dataset} & \multirow{2}{*}{Method} & \multicolumn{6}{c}{Layers}  \\
& & 2     & 4     & 8     & 16    & 32    & 64       \\

\midrule
\multirow{2}{*}{\begin{tabular}[c]{@{}l@{}}NTU-\\2012\end{tabular}} 
& HGNN & \textbf{83.16} & 77.59 & 5.36 & 5.20 & 5.28 & 5.25    \\
& Deep-HGCN & \textbf{84.50}& 84.32 & 83.87 & 82.60 & 81.61 & 80.42    \\
\midrule
\multirow{2}{*}{\begin{tabular}[c]{@{}l@{}}Model-\\Net40\end{tabular}}   
& HGNN & \textbf{95.96} & 94.71 & 4.31 & 4.27 & 4.14 & 4.13      \\
& Deep-HGCN & \textbf{96.08} & 96.00 & 94.54 &95.22   &95.20 & 88.24  \\
\bottomrule
\end{tabular}
}
    \vspace{-0mm}
\label{tab:object_accuracy}
\end{table}

Additional experiments results on the datasets given by HNHN~\cite{hnhn} can be seen in the Appendix Table \ref{tab:hnhn_accuracy}.

\section{Conclusions}
In this paper, We theoretically analyzed the over-smoothing problem of hypergraph convolutional neural networks from two perspectives: random walk on hyergraph and hypergraph Dirichlet energy. Specifically, the stationary distribution of random walk and the Dirichlet energy of hypernode embedding converging to zero both leads to over-smoothing. Therefore, we proposed a Deep-HGCN model with initial residual connection and identity mapping to avoid the over-smoothing issue. Furthermore, we gave theoretical justifications for Deep-HGCN's effectiveness against over-smoothing by proving that a $k$-layer Deep-HGCN simulates a polynomial filter of order $k$ with arbitrary coefficients.
Experiments show that the proposed model relieves the problem of over-smoothing and achieves state-of-the-art performance on the hypernode classification task.

\bibliographystyle{named}
\bibliography{ijcai21}
\newpage
\clearpage
\onecolumn
\appendix
    \begin{center}
    \Large
    \textbf{Appendix}
     \\[20pt]
    \end{center}

\section{Missing Proofs in Section \ref{sec:3.2}}
\subsection{The proof of Lemma~\ref{yl:sd1}}
\begin{align}
    \tilde{\pi}\P &=\mbf{1}^T\D_v^{\frac{1}{2}} \D_v^{-\frac{1}{2}} \H\W \D_v^{-\frac{1}{2}} \H^T \D_v^{-\frac{1}{2}} \notag  \\
    &=(\mbf{1}^T\D)(\D^{-1}\H\W \D_v^{-\frac{1}{2}} \H^T ) \D_v^{-\frac{1}{2}}
\end{align}
From Eq.\eqref{eq:sd1}, we know $ (\mbf{1}^T\mbf{D}_v)(\mbf{D}_{v}^{-1}\mbf{H}\mbf{W}\mbf{D}_e^{-1}\mbf{H}^T)=\mbf{1}^T\mbf{D}_v$, then we have 
\begin{align}
     \tilde{\pi}\P=  \mbf{1}^T\mbf{D}_v\D_v^{-\frac{1}{2}}=\mbf{1}^T\mbf{D}_v^{\frac{1}{2}}=\tilde{\pi}
\end{align}

\subsection{the proof of Theorem \ref{dl:1}.}
\textit{Proof}
According to \cite{zhou2007learning}, we can drive that the random walk defined on hypergraph have a unique stationary distribution ( proof by contradiction). Based on the Lemma ~\ref{yl:sd1}, we know the random walk with transition matrix $\P$ has unique stationary distribution. 

Let stationary distribution $\pi\in\R^{|\V|\times 1}$ and we have $\P\Tilde{\pi}=\Tilde{\pi}$,
\begin{align}
        lim_{k\to\infty}\P^{k}= \mbf Q=[\Tilde\pi,\Tilde{\pi},...,\Tilde{\pi}]
\end{align}
where $\Q\in \R^{|\V|\xx|\V|}$,k and the each column in $\Q$ equal to $\tilde{\pi}$. Then, it can derive that 
\begin{align}
    lim_{k\to \infty}\P^{k}\x_{*j}=\Q^T
    \x_{*,j}=\sum_{i=1}^{|\V|}\x_{ij}\tilde{\pi}
\end{align}

\section{Missing Proofs in Section \ref{sec:3.3}}

\begin{proof}[Lemma ~\ref{yl:2}]
Let us denote the eigenvalues of $\Delta$ by $\lambda_1, \lambda_2, \dots, \lambda_{|\V|}$, and the associated eigenvectors by $v_1, v_2, \dots, v_{|\V|}$.\\
Suppose $f=\sum_{i=1}^{|\V|} c_i v_i$ where $c_i\in \mathbb{R}$
\begin{equation}
    E(f)=f^T \Delta f=f^T \sum_{i=1}^{|\V|} c_i \lambda_i v_i=\sum_{i=1}^{|\V|} c_i^2 \lambda_i
\end{equation}
Therefore
\begin{equation*}
    \begin{aligned}
    E(Pf)&=f^T(I_{|\V|}-\Delta)^T\Delta (I_{|\V|}-\Delta)f
    =f^T(I_{|\V|}-\Delta) \Delta (I_{|\V|}-\Delta)f\\
    &=\sum_{i=1}^{|\V|} c_i^2 \lambda_i (1-\lambda_i)^2
    \le(1-\lambda)^2E(f)
    \end{aligned}
\end{equation*}\\
(2)
\begin{equation}
    \begin{aligned}
    E(\X\Th)&=tr(\Th^T\X^T\Delta\X\Th)
    =tr(\X^T\Delta\X\Th\Th^T)\\
    &\le \|\Th^T\|_2^2 tr(\X^T\Delta\X)
    =\|\Th^T\|_2^2E(\X)
    \end{aligned}
\end{equation}
(3)
\begin{align}
    		E(f)=f^T\Delta f
    		= \frac{1}{2}\sum_{e\in\E}\sum_{\{u,v\}\in \V}
    		\frac{w(e)h(u,e)h(v,e)}{\delta(e)}
    		&\left(\frac{f(u)}{\sqrt{d(u)}}-\frac{f(v)}{\sqrt{d(v)}}\right)^2,
    		\label{eq:DE}
    \end{align}
\begin{equation}
    \begin{aligned}
    	E(f)&=f^T\Delta f\\
    	&= \frac{1}{2}\sum_{e\in\E}\sum_{\{u,v\}\in \V}
    	\frac{w(e)h(u,e)h(v,e)}{\delta(e)} 
    	\left(\frac{f(u)}{\sqrt{d(u)}}-\frac{f(v)}{\sqrt{d(v)}}\right)^2\\
    	&\ge\frac{1}{2}\sum_{e\in\E}\sum_{\{u,v\}\in \V}
    	\frac{w(e)h(u,e)h(v,e)}{\delta(e)} 
    	\left(\sigma\left(\frac{f(u)}{\sqrt{d(u)}}\right)-\sigma\left(\frac{f(v)}{\sqrt{d(v)}}\right)\right)^2\\
    	&=\frac{1}{2}\sum_{e\in\E}\sum_{\{u,v\}\in \V}
    	\frac{w(e)h(u,e)h(v,e)}{\delta(e)} 
    	\left(\frac{\sigma(f(u))}{\sqrt{d(u)}}-\frac{\sigma(f(v))}{\sqrt{d(v)}}\right)^2\\
    	&=E(\sigma(f))
    \end{aligned}
\end{equation}
Then extending the above argument to vector field completes the proof.
\end{proof}

\begin{proof}[Theorem \ref{dl:2} ]
\begin{equation}
    \begin{aligned}
      E(\X^{(l)})&=E(\sigma(\P\X^{(l-1)}\Th^{(l-1))})
      \le E(\P\X^{(l-1)}\Th^{(l-1)})\\
      &\le \|\Th^{(l-1)}\|_2^2 E(\P\X^{(l-1)})
     =s_l E(\P\X^{l-1})\\
      &\le s_l\Bar{\lambda}E(\X^{(l-1)}) 
    \end{aligned}
\end{equation}
\end{proof}

\section{Missing Proofs in Section \ref{sec:4.2}}
\subsection{The proof of Theorem \ref{dl:3}}
{\it Proof.}
Taking the vertex features as graph signals, e.g., a column of the feature matrix $\X $ can be considered as a graph
signal.
For simplicity, we assume the signal vector $\x\in \R^{d}$ to be non-negative. Note that we can convert $\x$ into a non-negative input layer $\mbf{x}^{(0)}$ by a linear transformation $\mbf{x}^{(0)}=\mbf{\Th}\x\in\R^{c}$. We consider a weaker version of Deep-HGCN by fixing $\alpha_l=0.5$ and fixing the weight matrix $(1-\beta_l)\mbf{I}_c+\beta_l\Th^{(l)}$ to be $\gamma_l\mbf{I}_c$, where $\gamma_l$ is a learnable parameter. We have
\begin{equation}
  \begin{aligned}
  \mbf{x}^{(l+1)}=\frac{1}{2}\sigma\left(\mbf P(\mbf{x}^{(l)}+\x)\gamma_l\mbf{I}_c\right)
  \end{aligned}
\end{equation}
Since the input feature x is non-negative, we can remove coefficient $\frac{1}{2}$ and ReLU operation for simplicity:
k\\l\begin{equation}
  \begin{aligned}
   \mbf{x}^{(l+1)}=&\gamma_lP(\mbf{x}^{(l)}+\x)
&=\gamma_l\left((\mbf{I}_c-\Delta)(\mbf{x}^{(l)}+x)\right)
  \end{aligned}
\end{equation}
Consequently, we can express the final representation after $K$ layers Deep-HGCN as:
\begin{equation}
  \begin{aligned}
    \mbf{x}^{(l+1)}=\left(\sum_{l=0}^{K-1}\left(\prod_{k=K-l-1}^{K-1} \gamma_{k}\right)\left(\mbf{I}_c-\Delta\right)^l\right)\x.
  \end{aligned}
\end{equation}
On the other hand, as introduced in the begining, a polynomial filter of $K-$layer Deep-HGCN can be expressed as:
\begin{equation}
  \begin{aligned}
        \left(\sum_{k=0}^{K-1}\theta_{k} \Delta^{k}\right)x=\left(\sum_{k=0}^{K-1}\theta_k \left(\mbf{I}_{c}-\left(\mbf{I}_{c}-\Delta\right)
        \right)^k\right)\x\\
        =\left(\sum_{k=0}^{K-1}\theta_k \left(\sum_{l=0}^{k}(-1)^{l}\left(\begin{matrix}
        k\\l \end{matrix}\right)\left(\mbf{I}_c-\Delta\right)^{l}\right)\right)\x\\
        =\left(\sum_{l=0}^{K-1} \left(\sum_{k=l}^{K-1}\theta_k(-1)^{l}\left(\begin{matrix}
        k\\l \end{matrix}\right)\left(\mbf{I}_c-\Delta\right)^{l}\right)\right)\x
  \end{aligned}
\end{equation}

To show that Deep-HGCN can express an arbitrary $K$-order polynomial filter, we need to prove that there exists a solution $\gamma_l$, $l$=0, $\cdots$,$K-1$ such that the corresponding coefficients of $\left(\mbf{I}_c-\Delta\right)^{l}$ in (13) and (14) are equivalent. More precisely, we need to show the following equation system has a solution $\gamma_l$, $l$=0, $\cdots$,$K-1$.
\begin{equation}
    \prod_{k=K-l-1}^{K-1}\gamma_{k} =\sum_{l=0}^{k} \theta_{k}(-1)^{l}\left(\begin{matrix}
        k\\l \end{matrix}\right),k=0, \cdots,K-1.
\end{equation}

Note that we can solve the equation system by
\begin{equation}
  \begin{aligned}
   \gamma_{K-l-1}=\sum_{k=l}^{K-1} \theta_{k}(-1)^{l}\left(\begin{matrix}
        k\\l \end{matrix}\right)/\sum_{k=l-1}^{K-1} \theta_{k}(-1)^{l-1}\left(\begin{matrix}
        k\\l-1 \end{matrix}\right)
  \end{aligned}
\end{equation}
for $l$=0, $\cdots$,$K-2$ and $\gamma_{K-1l}=\sum_{k=0}^{K-1} \theta_{k}$.


\section{Missing Experiments in Section 5.2}
\subsection{Accuracy results}
\paragraph{Visual object classification.} See Table \ref{tab:object_accuracy}.
\paragraph{Citation network classificaion from HNHN.} See Table \ref{tab:hnhn_accuracy}.
\begin{table*}[htbp]
\def\p{$\pm$} 
\centering
\caption{Summary of classification accuracy(\%) results with various depths on object datasets.}
    \vspace{-2mm}
\scalebox{0.9}{
\begin{tabular}{cc|cccccc}
\toprule
\multirow{2}{*}{Dataset} & \multirow{2}{*}{Method} & \multicolumn{6}{c}{Layers}  \\
& & 2     & 4     & 8     & 16    & 32    & 64       \\

\midrule
\multirow{2}{*}{\begin{tabular}[c]{@{}l@{}}NTU2012\end{tabular}} 
& HGNN & \textbf{83.16\p0.6} & 77.59\p2.9 & 5.36\p0.2 & 5.20\p0.2 & 5.28\p0.2 & 5.25\p0.2    \\
& Deep-HGCN & \textbf{84.50\p0.5} & 84.32\p1.0 & 83.87\p0.6 & 82.60\p0.8 & 81.61\p0.9 & 80.42\p1.1    \\
\midrule
\multirow{2}{*}{\begin{tabular}[c]{@{}l@{}}ModelNet40\end{tabular}}   
& HGNN & \textbf{95.96\p0.5} & 94.71\p0.9 & 4.31\p0.2 & 4.27\p5.1 & 4.14\p0.1 & 4.13\p0.1      \\
& Deep-HGCN & \textbf{96.08\p0.4 } & 96.00\p0.5 & 94.54\p2.3& 95.22\p0.9 &  95.20\p0.5 & 88.24\p15  \\
\bottomrule
\end{tabular}
}
\label{tab:object_accuracy}
\end{table*}

\begin{table*}[htbp]
\def\p{$\pm$} 
\centering
\caption{Summary of classification accuracy(\%) results of the experiment on the datasets that provide by HNHN}
\scalebox{0.9}{
\begin{tabular}{cc|cccccc}
\toprule
\multirow{2}{*}{Dataset} & \multirow{2}{*}{Method} & \multicolumn{6}{c}{Layers}  \\
& & 2     & 4     & 8     & 16    & 32    & 64       \\

\midrule
\multirow{4}{*}{\begin{tabular}[c]{@{}l@{}}Cora\end{tabular}} 
& HGNN  & \textbf{57.67\p0.9} & 51.87\p3.1 & 42.64\p2.8  & 40.37\p0.05  & 40.38\p0.07   & 40.42\p0.6      \\
& HNHN~\cite{hnhn} & \textbf{55.06\p4.1} & 35.15\p2.7 & 34.54\p6.1 & 15.56\p10.9 & 4.54\p0.7 & 10.70\p3.5    \\
& Deep-HGCN & 58.89\p0.7& 58.92\p0.4 & 59.45\p0.3 & 57.60\p0.9 & \textbf{61.19\p0.3} & 60.97\p0.7    \\
\midrule
\multirow{4}{*}{\begin{tabular}[c]{@{}l@{}}Citeseer\end{tabular}}   
& HGNN  & \textbf{65.67\p0.7} & 63.63\p2.2 & 37.28\p12.7  &21.76\p0.2   &21.45\p0.5    &21.80\p0.3       \\
& HNHN & \textbf{64.68\p5.2} & 43.66\p5.7 & 27.64\p2.4 & 26.27\p4.6 & 17.94\p4.2 & 17.94\p4.2      \\
& Deep-HGCN & 65.72\p0.9& 67.62\p0.9 & 66.50\p1.4 & 67.09\p1.3  & 67.53\p1.1 & \textbf{67.89\p0.5}    \\
\bottomrule
\end{tabular}
}
\label{tab:hnhn_accuracy}
\end{table*}

\subsection{Datasets}
See Table \ref{tab:dataset_hypergcn},\ref{tab:dataset_object}, \ref{tab:dataset_hnhn}.
\begin{table}[htbp]
\centering
\caption{Real-world hypergraph datasets used in our work.(co-a represents co-authorship, and co-c represent co-citation)}
    \vspace{-2mm}
 \scalebox{0.88}{
\begin{tabular}{l|llll}
\toprule
Dataset& Hypernodes & Hyperedges & Features & Classes  \\
\midrule
\begin{tabular}[c]{@{}l@{}}\textbf{Cora}(co-a)\end{tabular}   
& 2708         & 1072         &1433   & 7          \\

\begin{tabular}[c]{@{}l@{}}\textbf{DBLP}(co-a)\end{tabular}   
& 43413        & 22535        & 1425       & 6          \\

\begin{tabular}[c]{@{}l@{}}\textbf{Pubmed}(co-c)\end{tabular}   
& 19717        & 7963         & 500        & 3          \\

\begin{tabular}[c]{@{}l@{}}\textbf{Cora}(co-c)\end{tabular}     
& 2708         & 1579         & 1433       & 7          \\

\begin{tabular}[c]{@{}l@{}}\textbf{Citeseer}(co-c)\end{tabular} 
& 3312         & 1079         & 3703       & 6          \\
\bottomrule
\end{tabular}
}
\label{tab:dataset_hypergcn}
\end{table}

\begin{table}[htbp]
\centering
\caption{The detailed information of the ModelNet40 and NTU datasets.}
    \vspace{-2mm}
 \scalebox{0.88}{
\begin{tabular}{l|ll}
\toprule
Dataset& MOdelNet40 & NTU2012  \\
\midrule
Objects   
& 12311 & 2012         \\

MVCNN Feature  
&4096  & 4096   \\
GVCNN Feature  
&2048  & 2048         \\
Training node  
&9843  & 1639       \\
Testing node  
&2468 &  373       \\
Classes 
&40  & 67         \\

\bottomrule
\end{tabular}
}
\label{tab:dataset_object}
\end{table}

\begin{table}[htbp]
\centering
\caption{Datasets used in the experiment compared with HNHN.}
    \vspace{-2mm}
 \scalebox{0.88}{
\begin{tabular}{l|llll}
\toprule
Dataset& Hypernodes & Hyperedges & Features & Classes\\
\midrule
\textbf{Cora} & 16313        & 7389        &1000   & 10     \\
\textbf{Citeseer}& 1498         & 1107         & 3703       & 6          \\
\bottomrule
\end{tabular}
}
\label{tab:dataset_hnhn}
\end{table}

\subsection{Hyper-parameter search strategy}
See Table \ref{tab:hyper_parameter}.
\begin{table}[htbp]
\centering
\caption{Hyper-parameter search range for Deep-HGCN model.}
    \vspace{-2mm}
 \scalebox{0.88}{
\begin{tabular}{l|ll}
\toprule
Hyper-parameter & Range  \\
\midrule
$\alpha$   & \{0 0.1 0.2 0.3 0.4 0.5 0.6 0.7 0.8 0.9 1\} \\
$\lambda$ &\{0, 0.5, 1.0,1.5,2.0,2.5\}  \\
\bottomrule
\end{tabular}
}
\label{tab:hyper_parameter}
\end{table}

\end{document}